# Unsupervised learning of object landmarks by factorized spatial embeddings

James Thewlis
University of Oxford
jdt@robots.ox.ac.uk

Hakan Bilen
University of Oxford
University of Edinburgh
hbilen@robots.ox.ac.uk

Andrea Vedaldi
University of Oxford
vedaldi@robots.ox.ac.uk

## Abstract

*Learning automatically the structure of object categories remains an important open problem in computer vision. In this paper, we propose a novel unsupervised approach that can discover and learn landmarks in object categories, thus characterizing their structure. Our approach is based on factorizing image deformations, as induced by a viewpoint change or an object deformation, by learning a deep neural network that detects landmarks consistently with such visual effects. Furthermore, we show that the learned landmarks establish meaningful correspondences between different object instances in a category without having to impose this requirement explicitly. We assess the method qualitatively on a variety of object types, natural and man-made. We also show that our unsupervised landmarks are highly predictive of manually-annotated landmarks in face benchmark datasets, and can be used to regress these with a high degree of accuracy.*

## 1. Introduction

The appearance of objects in images depends strongly not only on their intrinsic properties such as shape and material, but also on accidental factors such as viewpoint and illumination. Thus, learning from images about objects as intrinsic physical entities is extremely difficult, particularly if no supervision is provided.

Despite these difficulties, the performance of object detection algorithms has been rising steadily, and deep neural networks now achieve excellent results on benchmarks such as PASCAL VOC [17] and Microsoft COCO [39]. Still, it is unclear whether these models conceptualise objects as intrinsic entities. Early object detectors such as HOG [13] and DPMs [18] were based on 2D templates applied in a translation and scale invariant manner to images. Recent detectors such as SSD [42] make this even more extreme and learn different templates (filters) for different scales and even different aspect ratios of objects. Hence, these models are likely to capture objects as image-based phenomena, representing them as a collection of weakly-related 2D pat-terns.

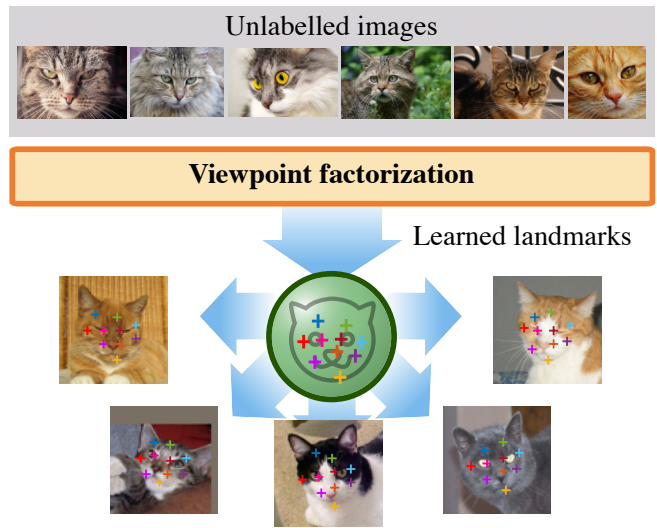

Figure 1. We present a novel method that can learn **viewpoint invariant landmarks without any supervision**. The method uses a process of viewpoint factorization which learns a deep landmark detector compatible with image deformations. It can be applied to rigid and deformable objects and object categories.

Achieving a deeper understanding of objects requires modeling their intrinsic viewpoint-independent structure. Often this structure is defined manually by specifying entities such as landmarks, parts, and skeletons. Given sufficient manual annotations, it is possible to teach deep neural networks and other models to recognize such structures in images. However, the problem of *learning such structures without manual supervision* remains largely open.

In this paper, we contribute a new approach to learn viewpoint-independent representations of objects from images without manual supervision (fig. 1). We formulate this task as a *factorization problem*, where the effects of image deformations, for example arising from a viewpoint change, are explained by the motion of a reference frame attached to the object and independent of the viewpoint.

After describing the general principle (sec. 3.1), we in-vestigate a particular instantiation of it. In this model, the structure of an object is expressed as a set of *landmark points* (sec. 3.2) detected by a neural network. Differently from traditional keypoint detectors, however, the network is learned without manual supervision. Learning considers pairs of images related by a warp and requires the detector's output to be *equivariant* with the transformation (sec. 3.3). Transformations could be induced by real-world viewpoint changes or object deformations, but we show that meaningful landmarks can be learned even by considering random perturbations only.

We show that this method works for individual rigid and deformable object *instances* (sec. 3.1.1) as well as for object *categories* (sec. 3.1.2). This only requires learning a single neural network to detect the same set of landmarks for images containing different object instances of a category. While there is no *explicit* constraint that forces landmarks for different instances to align, we show that, in practice, this tends to occur automatically.

The method is tested qualitatively on a variety of different object types, including shoes, animals, and human faces (sec. 4). We also show that the unsupervised landmarks are highly predictive of manually-annotated landmarks, and as such can be used to detect these with a high degree of accuracy. In this manner, our method can also be used for unsupervised pretraining of semantic landmark detectors.

## 2. Related work

**Flow.** Matching images up to a motion-induced deformation links back to the work of Horn and Schunck [26] on optical flow and to deep learning approaches for its computation [21, 57, 28]. Flow can also be defined semantically rather than geometrically [40, 32, 46, 77, 76]. While our method also establishes geometric and (indirectly) semantic correspondences, it goes beyond that by learning a single set of viewpoint independent landmarks which are valid for *all* images at once.

**Parts.** A traditional method to describe the structure of objects is to decompose them into their constituent parts. Several unsupervised methods to learn parts exist, from the constellation approach used in [19, 9, 62] to the Deformable Parts Model (DPM) [18] and many others. More recently, AnchorNet [48] successfully learns parts that match different object instances as well as different object categories using only image-level supervision; furthermore, they propose a part orthogonality constraint similar to our own. While the concepts of landmarks and parts are similar, our training method differs substantially from these approaches: rather than learning parts as a byproduct of learning a (deformable) discriminator, our landmark points are trained to fit geometric deformations directly.

**Deformation-prediction networks.** *WarpNet* [30] learns a neural network that, given two images, predicts a Thin Plate Spline (TPS [6]) that aligns them. While our landmarks can also be seen as a representation of transformations (as matching them between image pairs induces one), learning such landmarks is unique to our method. The Deep Deformation Network of [69] predicts image transformations to refine landmarks using a "Point Transformer Network", but their landmarks are learned using full manual supervision, whereas our method is fully unsupervised. Very recently [53] learn a neural network that also aligns two images by estimating the transformation between them, implicitly learning feature extractors that could be similar to keypoints; however, our work explicitly trains a network to output keypoints that are equivariant to such transformations.

**Landmark detection.** There is an extensive literature on landmark detectors, particularly for faces. Examples include Active Appearance Models [11], along with subsequent improvements [44, 12] and others using templates [51] or parts [80]. Other approaches directly regress the landmark coordinates [59, 14, 10, 52]. Deep learning methods use cascaded CNNs [56], coarse-to-fine autoencoders [70], auxiliary attribute prediction [73, 74], learned deformations [69] and LSTMs [64]. Beyond faces, there is work on humans [65, 58], birds [55, 41, 69] and furniture [63]. More general pose estimation including the case of landmarks is explored in [16]. Our method can build on any such detector architecture and can be used as a pretraining strategy to learn landmarks with less or no supervision.

**Equivariance constraint.** A variant of the equivariance constraint used by our method was proposed by [37] to learn feature point detectors for image matching. We build on a similar principle, but use it to learn intrinsic landmarks for object categories instead of generic SIFT-like features with a robust learning objective and learn to detect a set of complementary landmarks rather than a single one at a time.

**Unsupervised pretraining.** Unsupervised pretraining has received significant interest with the popularization of data-hungry deep networks [5, 24, 23]. Unsupervised learning is based on training a network to solve auxiliary tasks, for which supervision can be obtained without manual annotations. The most common of such tasks is to *generate* the data (autoencoders [7, 4, 25]); or one can remove some information in images and train a network to reconstruct it (denoising [60], ordering patches [15, 47], inpainting [50], analyzing motion [1, 49, 61, 20, 45], and colorizing [71, 35]). Our method can be seen in this light as trying to undo a synthetic deformation applied to an image.

Our method is also related to unsupervised learning for faces, such as alignment based on a face model [78], learning meaningful descriptors [67, 22], and learning a part model [38]. Huang *et al*. [27] learn joint alignment of faces using deep features, and Jaiswal *et al*. [29] use clustering to discover head modes in order to refine manually-defined landmarks in an unsupervised manner, both using genera-tive principles. None of these methods learns landmarks from scratch.

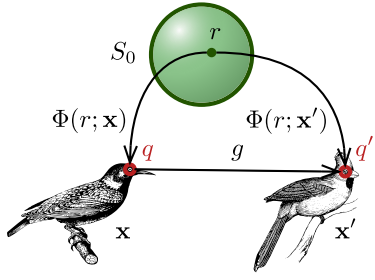

Figure 2. **Modelling the structure of objects.** Points $r$ in the reference space $S_0$ (conceptually a sphere) index corresponding points in different object instances. Given an image $\mathbf{x}$, the map $\Phi(r;\mathbf{x})$ detects the location $q$ of the reference point $r$. The map must be compatible with warps $g$ of the objects. For different views of the same (deformable) object instance, the warp $g$ is defined geometrically, whereas for object categories (as shown) it is defined semantically.

## 3. Method

Sec. 3.1 introduces the method of viewpoint factorization for learning an intrinsic reference frame for object instances and categories. Then, sec. 3.2 applies it to learn object landmarks and sec. 3.3 discusses the details of the learning formulation.

### 3.1. Structure from viewpoint factorization

Let $S \subset \mathbb{R}^3$ be the surface of a physical object, say a bird, and let $\mathbf{x} : \Lambda \to \mathbb{R}$ be an image of the object, where $\Lambda \subset \mathbb{R}^2$ is the image domain (fig. 2). The surface $S$ is an intrinsic property of the object, independent of the particular image $\mathbf{x}$ and of the corresponding viewpoint. We consider the problem of learning a function $q = \Phi_S(p; \mathbf{x})$ that maps object points $p \in S$ to the corresponding pixels $q \in \Lambda$ in the image.

We propose a new method to learn $\Phi_S$ automatically through a process of viewpoint factorization. To this end, consider a second image $\mathbf{x}'$ of the object seen from a different viewpoint. Occlusion not withstanding, one can write $\mathbf{x}' \approx \mathbf{x} \circ g$ where $g : \mathbb{R}^2 \to \mathbb{R}^2$ is the image warp induced by the viewpoint change. Using the map $\Phi_S$, the warp $g$ can be factorised as follows:

$$g = \Phi_S(\cdot; \mathbf{x}') \circ \Phi_S(\cdot; \mathbf{x})^{-1}. \qquad (1)$$

In other words, we can decompose the warp $g : q \mapsto q'$ as first finding the intrinsic object point $p = \Phi_S^{-1}(q; \mathbf{x})$ corresponding to pixel $q$ in image $\mathbf{x}$ and then finding the corresponding pixel $q' = \Phi_S(p; \mathbf{x}')$ in image $\mathbf{x}'$.

The factorization eq. (1) is more conveniently expressed as the following *equivariance constraint*:

$$\forall p \in S : \ \Phi_S(p; \mathbf{x} \circ g) = g(\Phi_S(p; \mathbf{x})). \qquad (2)$$

This constraint simply states that the points $p$ must be detected in a manner which is consistent with a viewpoint change.

In order to learn the map $\Phi_S$, we express the latter as a deep neural network and train it to satisfy constraint (2) in a Siamese configuration, supplying triplets $(\mathbf{x}, \mathbf{x}', g)$ to the learning process. Note that, if we are given two views $\mathbf{x}$ and $\mathbf{x}'$ of the same object, the viewpoint transformation $g$ is often unknown. Instead of trying to recover $g$, inspired by [30], we propose to synthesize transformations $g$ at random and use them to generate $\mathbf{x}'$ from $\mathbf{x}$. While this approach only uses unannotated images of the object, it can still learn meaningful landmarks (sec. 4).[1]

**Discussion.** While learning only considers deformations of the same image, the model still learns to bridge automatically across moderately different viewpoints (see fig. 5). However we leave very large out-of-plane rotations, which would require to handle partial occlusions of the landmarks, to future work.

#### 3.1.1 Deformable objects

The method developed above extends essentially with no modification to deformable objects. Suppose that the surface $S$ deforms between images according to isomorphisms $w : \mathbb{R}^3 \to \mathbb{R}^3$. We tie the shape variants $wS = \{w(p) : p \in S\}$ together by introducing a common reference space $S_0$, which we call an *object frame*. Barring topological changes, we can establish isomorphisms $\pi_S$ mapping reference points $r \in S_0$ to fixed surface points $\pi_S(r) \in S$, in the sense that $\forall w : w(\pi_S(r)) = \pi_{wS}(r)$. Then, by using the substitution $\Phi(r; \mathbf{x}) = \Phi_S(\pi_S(r); \mathbf{x})$, we can rewrite the equivariance constraint (2) as

$$\forall r \in S_0 : \ \Phi(r; \mathbf{x} \circ g) = g(\Phi(r; \mathbf{x})). \qquad (3)$$

This simply states that one expects surface points to be detected equivariantly with viewpoint-induced deformations as well as with deformations of the object surface.

#### 3.1.2 Object categories

In addition to deformable objects, our formulation can easily account for shape variations between object instances in the same category. To do this, one simply makes the assumption that all object surfaces $S$ are isomorphic to the same reference shape $S_0$ (fig. 2).

Differently from the case of deformable objects, geometry alone does not force the mappings $\pi_S$ for different object instances $S$ to be related. Nevertheless, we would like to choose such mappings to be *semantically consistent*; for example, if $\pi_S(r)$ is the right eye of face $S$, then we would like $\pi_{S'}(r)$ to be the right eye of face $S'$. An important contribution of this work is to show that semantically-meaningful correspondences emerge automatically by simply sharing *the same learned mapping* $\Phi$ between all object instances in a given category. The idea is that, by learning a single rule that detects object points consistently with deformations, these points tend to align between different object instances as this is the smoothest solution.

---

[1]If $\mathbf{x}$ and $\mathbf{x}'$ are given but $g$ is unknown, one can rewrite eq. (2) by expressing the warp $g$ as a function of the predicted landmarks (as the solution of the equation $\forall p : \Phi_S(p; \mathbf{x}') = g\Phi_S(p; \mathbf{x})$), and then by measuring the alignment quality in appearance space as $\|\mathbf{x}' - \mathbf{x} \circ g\|$. However, this approach provides a weaker supervisory signal and is somewhat more complex to implement.

## 3.2. Landmark detection networks

In this section we instantiate concretely the method of sec. 3.1. First, one needs to decide how to represent the maps $\Phi(\cdot; \mathbf{x}) : S_0 \to \Lambda$ as the output of a neural network or other computational model. Our approach is to *sample* this function at a set of $K$ discrete reference locations $\Phi(\mathbf{x}) = (\Phi(r_1; \mathbf{x}), \ldots, \Phi(r_K; \mathbf{x}))$. In this manner, the function $\Phi(\mathbf{x})$ can be thought of as detecting the location $p_k = \Phi(r_k; \mathbf{x})$ of $K$ *object landmarks*. We do not attach particular constraints to the set of landmarks, which can be thought of as an index set $r_k = k, k = 1, 2, \ldots, K$.

If $\Phi$ is implemented as a neural network, one can use any of the existing architectures for keypoint detection (sec. 2). Most such architectures are based on estimating *score maps* $\Psi(\mathbf{x}) \in \mathbb{R}^{H \times W \times K}$, associating a score $\Psi(\mathbf{x})_{uk}$ to each landmark $r_k$ and image location $u \in \{1, \ldots, H\} \times \{1, \ldots, W\} \subset \mathbb{R}^2$. The score maps can be transformed into probability maps by using the *softmax* operator $\sigma$:

$$p(u|\mathbf{x}, r) = \sigma[\Psi(\mathbf{x})]_{ur} = \frac{e^{\Psi(\mathbf{x})_{ur}}}{\sum_v e^{\Psi(\mathbf{x})_{vr}}}.$$

Following [66], it is then possible to extract a landmark location by using the soft argmax operator, which computes the expected value of this density:

$$u_r^* = \sigma_{\arg}[\Psi(\mathbf{x})]_r = \sum_u u\, p(u|\mathbf{x}, r) = \frac{\sum_u u e^{\Psi(\mathbf{x})_{ur}}}{\sum_v e^{\Psi(\mathbf{x})_{vr}}}.$$

The overall network, computing the location of the $K$ landmarks, can then be expressed as

$$\Phi(\mathbf{x}) = \sigma_{\arg}[\Psi(\mathbf{x})]. \quad (4)$$

**Discussion.** An alternative approach for representing the maps $S_0 \to \Lambda$ is to predict the parameters of a parametric transformation $t$. Assuming that the reference set $S_0 \subset \mathbb{R}^2$ is a space of continuous coordinates, the transformation $t$ could be an affine one [37] or a thin plate spline (TPS) [30]. This has the advantage of capturing in one step a dense set of object points and can be used to impose smoothness on the map.

However, using discrete landmarks is more robust and general. For example, individual landmarks may be undetectable when occluded, and this model can handle this case more easily without disrupting the estimate of the visible landmarks. Furthermore, one does not need to make assumptions on the family of allowable transformations, which could be difficult in general.

## 3.3. Learning formulation

In this section, we show how the equivariance constraint (3) can be used to learn $\Phi$ from examples. The idea is to setup the learning problem as a *Siamese* configuration, in which the output of $\Phi$ on two images $\mathbf{x}$ and $\mathbf{x}'$ is assessed for compatibility with respect to the deformation $g$ and the equivariance constraint (3). We can express this condition as the loss term:

$$\mathcal{L}_{\text{align}} = \frac{1}{K} \sum_{r=1}^{K} \|\Phi(\mathbf{x} \circ g)_r - g(\Phi(\mathbf{x})_r)\|^2. \quad (5)$$

In the rest of the section, we discuss two extensions to eq. (5) that allow the system to train better landmarks: formulating the loss directly in terms of the keypoint probabilities and adding a diversity term.

**Probability maps loss.** Equation (5) uses the soft argmax operator in order to localise and then compare landmarks. We show here that one can skip this step by writing a loss directly in terms of the probability maps, which provides a more direct and stable gradient signal. The idea is to replace eq. (5) with the loss term

$$\mathcal{L}'_{\text{align}} = \frac{1}{K} \sum_{r=1}^{K} \sum_{uv} \|u - g(v)\|^2 p(u|\mathbf{x}, r) p(v|\mathbf{x}', r) \quad (6)$$

where $p(u|\mathbf{x}, r) = \sigma[\Psi(\mathbf{x})]_{ur}$ and $p(v|\mathbf{x}', r) = \sigma[\Psi(\mathbf{x}')]_{vr}$ are the landmark probability maps extracted from images $\mathbf{x}$ and $\mathbf{x}'$.

Minimizing loss (6) has two desirable effects. First, it encourages the two probability maps to overlap and, second, it encourages them to be highly concentrated. In fact, the loss is zero if, and only if, both $p$ and $q$ are delta functions *and* if the corresponding landmark locations match up to $g$.

While a naive implementation of (6) requires to visit all pairs of pixels $u$ and $v$ in both images, with a quadratic complexity, a linear-time implementation is possible by decomposing the loss as:

$$\sum_u \|u\|^2 p(u|\mathbf{x}, r) + \sum_v \|g(v)\|^2 p(v|\mathbf{x}', r)$$
$$- 2 \left( \sum_u u\, p(u|\mathbf{x}, r) \right)^\top \cdot \left( \sum_v g(v) p(v|\mathbf{x}', r) \right).$$

**Diversity loss.** The equivariance constraint eq. (3) and its corresponding losses eqs. (5) and (6) ensure that the network learns at least one landmark aligned with image deformations. However, there is nothing to prevent the network from learning $K$ identical copies of the same landmark.

In order to avoid this degenerate solution, we add a *diversity* loss that requires probability maps of different landmarks to fire in different parts of the image. The most obvious approach is to penalize the mutual *overlap* between maps for different landmarks $r$ and $r'$:

$$\mathcal{L}_{\text{div}}(\mathbf{x}) = \frac{1}{K^2} \sum_{r=1}^{K} \sum_{r'=1}^{K} \sum_{u} p(u|\mathbf{x}, r) p(u|\mathbf{x}, r'). \quad (7)$$

This term is zero only if, and only if, the support of the different probability maps is disjoint.

The disadvantage of this approach is that it is *quadratic* in the number of landmarks. An alternative and more efficient diversity loss is:

$$\mathcal{L}'_{\text{div}}(\mathbf{x}) = \sum_{u} \left( \sum_{r=1}^{K} p(u|\mathbf{x}, r) - \max_{r=1,\dots,K} p(u|\mathbf{x}, r) \right). \quad (8)$$

Just like eq. (7), this loss is zero only if the support of the distributions is disjoint. In fact the sum of probability values at a given point $u$ is always greater than the max unless all but one probability are zero. Note that we can rewrite (8) more compactly as:

$$\mathcal{L}'_{\text{div}}(\mathbf{x}) = K - \sum_{u} \max_{r=1,\dots,K} p(u|\mathbf{x}, r).$$

In practice, we found it beneficial to apply the diversity loss after *downsampling* (by $m \times m$ sum pooling) the probability maps as this encourages landmarks to be extracted farther apart. Thus we consider:

$$\mathcal{L}''_{\text{div}}(\mathbf{x}) = K - \sum_{u} \max_{r=1,\dots,K} \sum_{\delta_u} p(mu + \delta_u|\mathbf{x}, r).$$

where $\delta_u \in \{0, \dots, m-1\}^2$.

**Learning objective.** The learning objective considers triplets $(\mathbf{x}_i, \mathbf{x}'_i, g_i)$ of images $\mathbf{x}_i$ and $\mathbf{x}'_i$ related by a viewpoint warp $g_i$ and optimizes:

$$\min_{\Psi} \lambda \mathcal{R}(\Psi) + \frac{1}{N} \sum_{i=1}^{N} \Big( \mathcal{L}'_{\text{align}}(\mathbf{x}_i, \mathbf{x}'_i, g_i; \Psi) + \gamma \mathcal{L}''_{\text{div}}(\mathbf{x}_i; \Psi) + \gamma \mathcal{L}''_{\text{div}}(\mathbf{x}'_i; \Psi) \Big), \quad (9)$$

where $\mathcal{R}$ is a regulariser (weight shrinkage for a neural network). As noted before, if triplets are not available, they can be *synthesized* by applying a random transformation $g_i$ to an image $\mathbf{x}_i$ to obtain $\mathbf{x}'_i = \mathbf{x}_i \circ g$. Note that all functions are easily differentiable for backpropagation.

## 4. Experiments

In this section, we first describe the implementation details (sec. 4.1) and then report both qualitative (sec. 4.2) and quantitative (sec. 4.3) results demonstrating the power of our unsupervised landmark learning method.

### 4.1. Implementation details

In all the experiments, the detector $\Phi$ contains six convolutional layers with 20, 48, 64, 80, 256, $K$ filters respectively, where $K$ is the number of object landmarks. Each convolutional layer is followed by a batch normalization and a ReLU layer. This network is proposed in [74] for supervised facial keypoint estimation. Differently, instead of downsampling the feature map after each convolutional layer, we use only one $2 \times 2$ max pooling layer with a stride of 2 after the first convolutional layer (conv1). Thus, given an input size of $H \times W \times 3$, the network outputs an $\frac{H}{2} \times \frac{W}{2} \times K$ feature map. We apply a spatial softmax operator to the output of the last convolutional layer to obtain $K$ probability maps, one for each landmark.

During training, we supply a set of triplets of $(\mathbf{x}_i, \mathbf{x}'_i, g_i)$ as input to the network. In order to generate them, given an example image $\mathbf{I}$, one can naively sample a random TPS and warp the image accordingly. However, as the input images are typically centered and at most very slightly rotated, the learned weights can be biased towards such a setting. Instead, we randomly sample two TPS transformations $(g_1, g_2)$ and consecutively warp the given image to generate an image pair *i.e.* $\mathbf{x} = \mathbf{I} \circ g_1$ and $\mathbf{x}' = \mathbf{x} \circ g_2$ (computed using inverse image warping as $\mathbf{x} \circ (g_2 \circ g_1)$). The TPS warps are parametrized as in [6] which can be decomposed into affine and deformation parts. To render realistic and diverse warps, we randomly sample scale, rotation angle and translation parameters within the pre-determined ranges. Examples of the transformations are shown in figs. 3 to 5.

We initialize the weights of convolutions with random gaussian noise and optimize the objective function (eq. (9)) (weight decay $\lambda = 5 \cdot 10^{-4}$, $\gamma = 500$) by using Adam [33] with an initial learning rate $10^{-4}$ until convergence, then reduce it by one tenth until no further improvement is seen.

### 4.2. Qualitative results

We train our unsupervised landmarks from scratch on three different domains: shoes (fig. 3), cat faces (fig. 4), and faces (fig. 5), and assess them qualitatively. We train landmark detectors on 49525 shoes from the UT Zappos50k dataset of [68] and 8609 images from the cat heads dataset of [72] and keep the rest for validation. Facial landmarks are learned on the CelebA dataset [43] which contains more than 200k celebrity images for 10k identities with 5 annotated landmarks. We use the provided cropped face images, which are roughly centered and scaled to the same size.

We train an 8 or 10 landmark network for each of the tasks to allow for clearer visualization. In addition, we show examples of a 30-landmark network for faces in fig. 6. In all cases we observe that: i) landmarks are detected consistently up to synthetic warps (affine or TPS) of the corresponding images and that ii) as a byproduct of learning to be consistent with such transformations, landmarks are very consistent across different object instances as well.

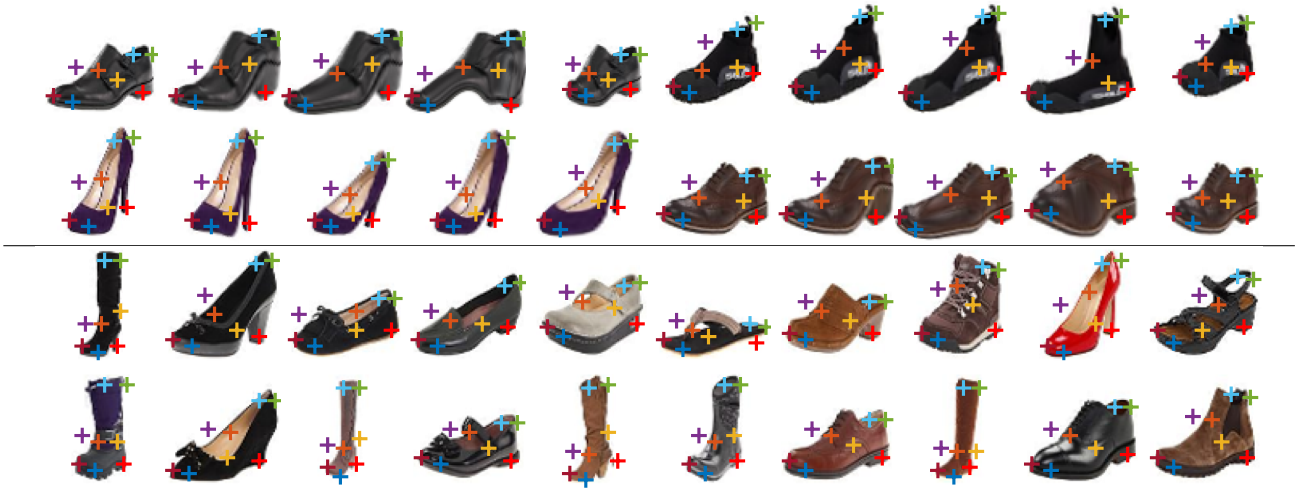

Figure 3. Unsupervised landmarks on shoes (8 landmark network). Top: synthetic TPS deformations (original image leftmost). Bottom: different instances. Note that landmarks are consistently detected despite the significant variation in pose, shape, materials, etc.

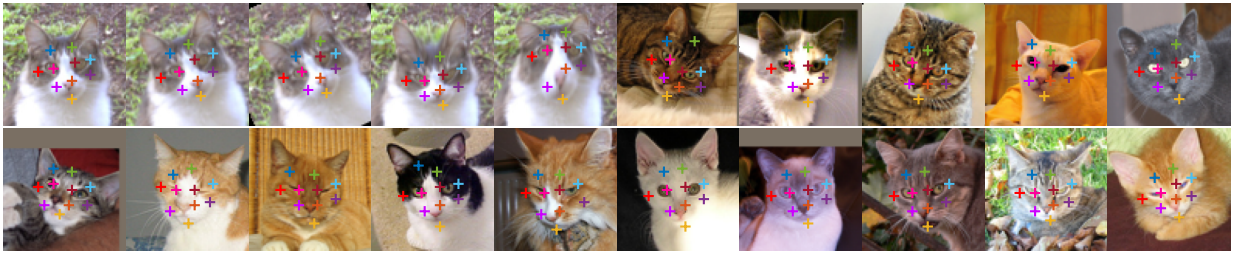

Figure 4. Unsupervised landmarks on cat faces (10 landmark network). Top-left quintuple: synthetic deformations (original image leftmost) transformed by rotation (images 2,3) and TPS warps (images 4,5). Remaining examples: different instances.

| $n$ landmarks | Regressor training | Mean error |
|---|---|---|
| 10 | MAFL | 7.95 |
| 30 | MAFL | 7.15 |
| 50 | MAFL | 6.67 |
| 10 | CelebA | 6.32 |
| 30 | CelebA | 5.76 |
| 50 | CelebA | **5.33** |

Table 1. Results on MAFL test set in terms of the inter-ocular distance as in [74, 52]. For each setting, $n$ unsupervised landmarks, that is learned on the CelebA training set, are regressed into 5 manually-defined landmarks. The regressor is learnt on CelebA or MAFL training set.

### 4.3. Quantitative results

In this section we evaluate the performance of our unsupervised landmarks quantitatively by testing how well they correlate with and predict manually-labelled landmarks. To do this, we consider standard facial landmark benchmarks containing manual annotations for semantic landmarks (e.g. eyes, corner of the mouth, etc). We first learn a detector for $K$ landmarks without supervision, freeze its weights, and then use the supervised training data in the benchmark to learn a linear regressor mapping the unsupervised landmark to the manually defined ones. The regressor takes as input the $2K$ coordinates of the unsupervised landmarks, stacks them in a vector $x \in \mathbb{R}^{2K}$, and maps the latter to the corresponding coordinates of the manually-defined landmarks as $y = Wx$. Learning $W$ can be seen as a fully connected layer with no bias, and is trained similarly to the unsupervised network, using our warps as data augmentation. Note that there is no backpropagation to the unsupervised weights, which remain fixed. $W$ is visualized in fig. 7.

| Method | Mean Error |
|---|---|
| TCDCN [74] | 7.95 |
| Cascaded CNN [56] | 9.73 |
| CFAN [70] | 15.84 |
| Our Method (50 points) | **6.67** |

Table 2. Comparison to state-of-the-art supervised landmark detectors on MAFL.

| Method | Mean Error |
|---|---|
| RCPR [8] | 11.6 |
| Cascaded CNN [56] | 8.97 |
| CFAN [70] | 10.94 |
| TCDCN [74] | 7.65 |
| RAR [64] | 7.23 |
| Our Method (51 points) | 10.53 |

Table 3. Comparison to state-of-the-art supervised landmark detectors on AFLW (5 pts) in terms of inter-ocular distance.

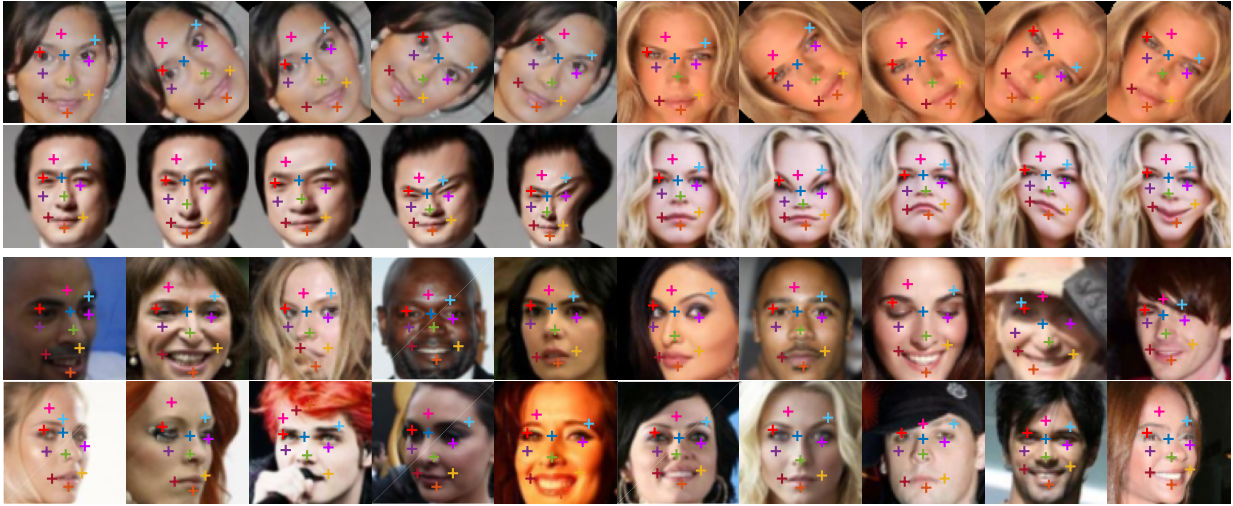

Figure 5. Unsupervised landmarks on CelebA faces (10 landmarks network). Top: synthetic rigid and TPS deformations (original image leftmost). Bottom: different instances. We observe landmarks highly aligned with facial features such as the mouth corners and eyes. Note that, being unsupervised, it needn't prefer the centers of the eyes, but consistently localizes points on the eye boundary.

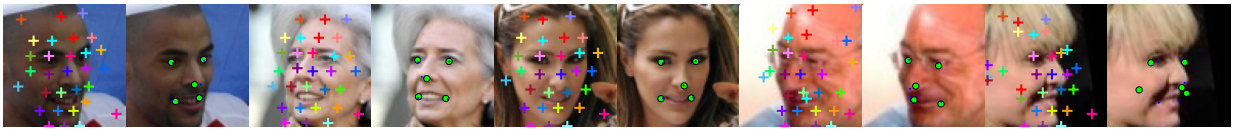

Figure 6. Regression of supervised landmarks form 30 unsupervised ones (left in each pair) on MAFL. The green dot is the predicted annotation and a small blue dot marks the ground-truth. A failure case is shown to the right.

| Supervised training images | Mean Error |
|---|---|
| All (19,000) | 7.15 |
| 20 | 8.06 |
| 10 | 8.49 |
| 5 | 9.25 |
| 1 | 10.82 |

Table 4. Localization results for different number of training images from MAFL used for supervised training.

| Method | Mean Error (68 pts) |
|---|---|
| DRMF [2] | 9.22 |
| CFAN [70] | 7.69 |
| ESR [10] | 7.58 |
| ERT [31] | 6.40 |
| LBF [52] | 6.32 |
| CFSS [79] | 5.76 |
| cGPRT [36] | 5.71 |
| DDN [69] | 5.65 |
| TCDCN [74] | 5.54 |
| RAR [64] | 4.94 |
| Ours (50 landmarks) | 9.30 |
| Ours (50 landmarks, finetune) | 7.97 |

Table 5. Comparison to state-of-the-art supervised landmark detectors on 300-W.

**Benchmark data.** We first report results on the MAFL dataset [74], a subset of CelebA with 19k training images and 1k test images annotated with 5 facial landmarks (corners of mouth, eyes and nose). We follow the standard evaluation procedure in [74] and report errors in inter-ocular distance (IOD) in table 1. Since the MAFL test set and the CelebA training set overlap partially, we remove the MAFL test images from CelebA when the latter is used for training.

We also consider the more challenging 300-W dataset [54] containing 68 landmarks, obtained by merging and re-annotating other benchmarks. We follow [52] and use 3148 images from AFW [80], LFPW-train [3] and Helen-train [75] as training set, and 689 images from IBUG, LFPW-test and Helen-test as test set.

Finally we use the AFLW [34] dataset, which contains

24,386 faces from Flickr. Although it contains up to 21 annotated landmarks, we follow [74, 64] in only evaluating five and testing on the same 2995 faces cropped and distributed in the MTFL set of [73]. For training we use 10,122 faces that have all five points labelled and whose images are not in the test set.

**MAFL results.** First, we train the unsupervised landmarks on the CelebA training set and learn a corresponding regressor on the MAFL training set. The accuracy of the regressor on the MAFL test data is reported in table 1 and qualitative results are shown in fig. 6.

Regressing from $K = 10, 30, 50$ unsupervised landmarks improves the results. This can be explained by the fact that more unsupervised landmarks means a higher chance of finding some highly correlated with the five manually-labelled ones and thus a more robust mapping (fig. 7). This can also increase accuracy since our landmarks are detected with a resolution of two pixels (due to the downsampling in the network). Table 2 compares these results to state-of-the-art *fully supervised* landmark localization methods. Encouragingly, our best regressor outperforms the supervised methods (6.67 error rate vs 7.95 of TCDCN [74]). This shows that our unsupervised training method is indeed able to find meaningful landmarks.

Next, in Table 4 we assess how many manual landmark annotations are required to learn the regressor. We consider the problem of regressing from $K = 30$ unsupervised landmarks and we observe that the regressor performs well even if only 10 or 20 images are considered (errors 8.5 and 8.06). By comparison, using all 19,000 training samples reduces the error to 7.15, which shows that most of the required information is contained in the unsupervised landmarks from the outset. This indicates that our method is very effective for **unsupervised pretraining** of manually annotated landmarks as well, and can be used to learn good semantic landmarks with few annotations.

**300-W results.** We use our best performing model, the 50 point network, trained unsupervised on CelebA, and report results in table 5 for two settings. In the first one, the unsupervised landmarks are learned on CelebA and only the regressor is learned on the 300-W training set; we obtain an error of 9.30. In the second setting, the unsupervised detector is fine-tuned (also without supervision) on the 300-W data to adapt the features to the target dataset. The fine-tuning lowers the error to 7.97 and yields a comparable result with the state-of-the-art supervised methods. This shows another strength of our method: our unsupervised learner can be used to adapt an existing network to new datasets, also without using labels.

**AFLW results.** Due to tighter face crops, we adapt our 50-landmark CelebA network, fine-tuning it first on similarly cropped CelebA images and then on the AFLW training set. The adapted network has 51 landmarks. We compare against other methods in table 3. Once more, landmarks linearly-regressed from the unsupervised ones are competitive with fully supervised detectors (10.53 vs 7.23). The regressor can be trained with as low as 1 or 5 labelled images almost saturating performance (errors 14.79 and 12.94 respectively). By comparison, the same architecture trained supervised from scratch using 5 and 10 labelled images with TPS data augmentation but no unsupervised pretraining has substantially higher 23.85 and 22.31 errors (achieved essentially by predicting the average landmark locations which has error 24.40).

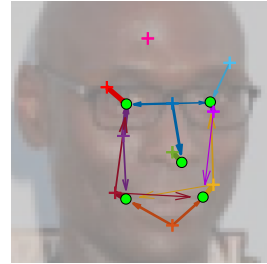

Figure 7. **Unsupervised ↔ supervised landmark correlation.** The thickness of each arrow from our unsupervised landmarks (crosses) to the supervised ones (circles) represents the averaged magnitude of each contribution in the learned linear regressor.

We also visualize what the regressor learns and which of the source (discovered) landmarks contribute to the target (semantic) ones in fig. 7. To do so, for each target landmark, we take the corresponding column of the regressor, compute the absolute value of its coefficients, $\ell^1$ normalize it, remove the entries smaller than $0.2$. We show this mapping as a directional graph with arrows between the target landmarks (green circular nodes) and the source ones (colored crosses). We observe that the contributions are proportional to the distance between source and target points. In addition, the landmark on the forehead, not in the convex hull of the target points is ignored, as expected.

## 5. Conclusions

In this paper we have presented a novel approach to learn the structure of objects in an *unsupervised manner*. Our key contribution is to reduce this problem to the one of learning landmark detectors that are equivariant, i.e. compatible, with image deformations. This can be seen as a particular instantiation of the more general idea of *factorizing deformations* by learning an intrinsic reference frame for the object. We have shown that this technique works for rigid and deformable objects as well as object categories, it results in landmarks highly-predictive of manually annotated ones, and can be used effectively for pretraining.

**Acknowledgments:** This work acknowledges the support of the AIMS CDT (EPSRC EP/L015897/1) and ERC 677195-IDIU.

# Supplementary Material: Unsupervised learning of object landmarks by factorized spatial embeddings

James Thewlis
University of Oxford
jdt@robots.ox.ac.uk

Hakan Bilen
University of Oxford
University of Edinburgh
hbilen@robots.ox.ac.uk

Andrea Vedaldi
University of Oxford
vedaldi@robots.ox.ac.uk

## 1. Introduction

In this supplementary material we elaborate on several details regarding the experimental setup, provide an additional comparison with training a supervised network on small numbers of images and present numerous images giving a qualitative look at the performance of our method. It is organized as follows: Sec. 2 gives the additional details and hyperparameters, Sec. 3 compares quantitatively with a supervised network and qualitative results are shown in Sec. 4. We evaluate the learned features on face segmentation in Sec. 5.

## 2. Experimental details

As described in Section 4.1 of the main text, we generate a pair of warps $(g_1, g_2)$. These are parameterized as Thin Plate Spline warps, which models the deformation of several keypoints along with an affine component. We sample all parameters from a gaussian with zero mean and the given standard deviations unless otherwise stated. The source keypoints are a $10 \times 10$ regular grid ($5 \times 5$ for MNIST), whereas each element of the parameter vector defining the destination keypoints is sampled with standard deviation $\sigma_{g_i,w}$. For each element we then add with 50% probability an additional perturbation sampled with standard deviation $\sigma_{g_i,W}$.

The affine component is parameterised as a similarity transform with rotation standard deviation $\sigma_{g_i,r}$ degrees, translation $\sigma_{g_i,t}$, and scale $\sigma_{g_i,s}$ with mean 1. Note we operate with normalized coordinated in the range $[-1, 1]$. Values are shown in Table 1. For faces and cats the input image dimensions are $100 \times 100$, which are then cropped after warping to $80 \times 80$. For MNIST the input images are resized to $35 \times 35$ then padded with a 5 pixel black border to be $45 \times 45$. For shoes the $64 \times 64$ initial images are padded with a 15 pixel white border to be $94 \times 94$.

The pooling layer prior to the diversity loss has pooling window size $5 \times 5$ in all networks except for MNIST and the AFLW 51 landmark network which have $3 \times 3$ (resulting in denser coverage of the face area, fig. 8).

|  | $g_i$ | $\sigma_{g_i,w}$ | $\sigma_{g_i,W}$ | $\sigma_{g_i,r}$ | $\sigma_{g_i,t}$ | $\sigma_{g_i,s}$ |
|---|---|---|---|---|---|---|
| Faces | $g_1$ | 0.001 | 0.001 | 0° | 0 | 0 |
|  | $g_2$ | 0.001 | 0.01 | 20° | 0.1 | 0.05 |
| MNIST | $g_1$ | 0.005 | 0.01 | 15° | 0.1 | 0.05 |
|  | $g_2$ | 0.005 | 0.02 | 20° | 0.1 | 0.05 |

Table 1. Standard deviations used for sampling warp parameters.

| Labelled Images | Sup. Net | Unsup. + Regressor |
|---|---|---|
| CelebA + AFLW | 8.67 | — |
| AFLW (10,122) | 14.25 | 10.53 |
| 20 | 21.13 | 13.28 |
| 10 | 22.31 | 13.85 |
| 5 | 23.85 | 12.94 |
| 1 | 28.87 | 14.79 |

Table 2. Results on AFLW (2995 images, 5 landmarks), varying the number of images used to train both a supervised network from scratch and a regressor on top of our unsupervised landmarks.

## 3. Supervised Network Comparison

In order to further evaluate the advantage of our unsupervised pre-training when a limited number of labelled images are used for subsequent supervised training, we compare to training a supervised network from scratch on the same images (Table 2 and fig. 1). The results reported in the main text adapted our unsupervised architecture with the addition of a final pooling layer (stride 2) and fully connected layer, achieving 23.85 error for 20 images. Here we train a network more comparable to existing supervised landmark networks by including pooling layers (stride 2) after the first three convolutional layers and taking a $64 \times 64$ input. It achieves results comparable to existing work when trained on many images and evaluated on AFLW (8.67, compare to Table 3 in the main paper) and error of 21.13 on 20 images. This confirms the advantage of our approach in the case of limited labelled data.

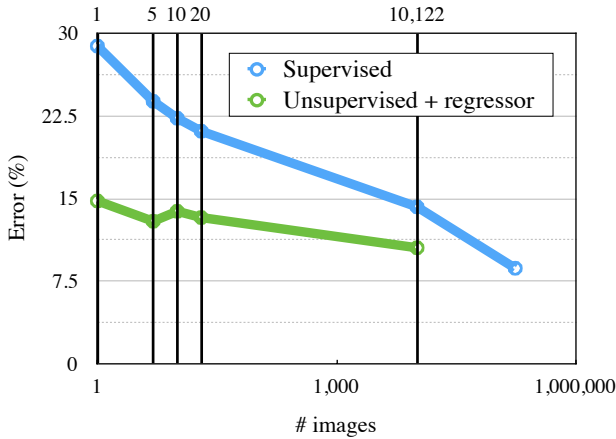

Figure 1. The same data as table 2 in graphical format.

## 4. Qualitative Results

We show additional images displaying the results of our method on different datasets and with different numbers of unsupervised landmarks.

The MNIST dataset of handwritten digits provides a simple setting in which to demonstrate the ability of our approach to identify landmarks across variations in writing style. We train separate networks for the digits 3, 5, and 6. The training data is augmented with Thin Plate Spline transformations and similarity transforms (parameters in Table 1). For each digit we use 1000 images for validation and the rest (around 5000) for training. As shown in fig. 3 the discovered landmarks are robust to rotations and significant differences in style.

To complement the examples of a 10-landmark network on cat faces in the main paper, we also show a network with 20 landmarks (fig. 4).

For the CelebA faces dataset (MAFL test subset) we show examples of a 30-landmark network (fig. 5) and the results of training our regressor with varying numbers of landmarks (fig. 7). For the 300-W dataset we show regression examples for a 30-landmark network (fig. 6). We also show the result of the 51-landmark network finetuned on AFLW and the regressor predictions (fig. 8).

In order to evaluate the effectiveness of our network in cases of illumination variation, we apply our 10-landmark CelebA network on frontal faces from the Cropped Extended Yale B[1] dataset. This dataset represents a significantly different domain to that used for training, being grayscale and tightly cropped. Nevertheless, with the more moderate lighting variations we get consistent landmarks. Failure occurs in the cases of hard shadows where there is little resolvable detail in areas of the face. We can fix this failure by finetuning to the target dataset, whereupon landmarks are predicted well across illumination variants. This finetuning can be done using synthetic warps as in the main paper, however we also note that in the cases of datasets like Yale B which offer aligned pairs of the same subject, we can simply train based on the identity transformation between aligned images having different lighting conditions. Both methods give qualitatively good results as shown in fig. 2.

[1]Georghiades, A. S., Belhumeur, P. N., & Kriegman, D. J. From few to many: Illumination cone models for face recognition under variable lighting and pose. PAMI 2001
Lee, K. C., Ho, J., & Kriegman, D. J. Acquiring linear subspaces for face recognition under variable lighting. PAMI 2005

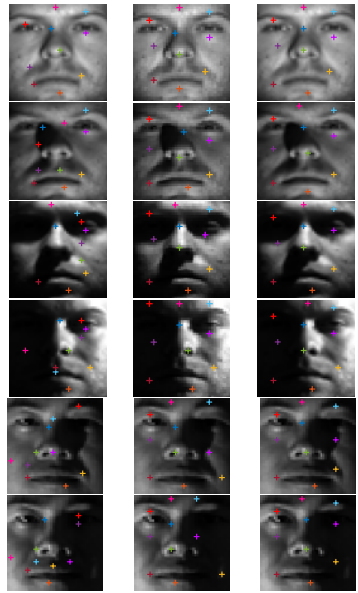

Figure 2. YaleB: Predicted landmarks on two Yale B subjects (held out during finetuning). Column 1: Original CelebA network, with poor results in shadows. Column 2: Finetuning from synthetic warps. Column 3: Finetuning from pairs with different lighting conditions.

|  | Epoch 5 | Epoch 30 | Best |
|---|---|---|---|
| From Scratch | 78.10% | 86.15% | 93.59% |
| Pretrained | 90.64% | 92.12% | 94.46% |
| Pretrained+ft | 90.84% | 92.41% | 94.82% |

Table 3. Pixel accuracy on HELEN when training from scratch, pretraining using our method (Conv 1-3 frozen) and pretraining while finetuning all layers

|  | 20 Images | 50 Images |
|---|---|---|
| From Scratch | 86.52% | 86.91% |
| Pretrained | 90.24% | 90.63% |

Table 4. Pixel accuracy on HELEN segmentation for a limited number of training images.

## 5. Evaluating learned features

We would like to know if the features obtained using our method are useful for other tasks. For this we use the task of face segmentation using the HELEN[2] dataset. We resize the images but do not further crop or preprocess them. We use our pretrained 50-landmark CelebA network with the first three layers frozen and replace the last layer with a 10-way spatial classification. We get 94.46% pixel accuracy, compared to 93.59% for the same network configuration trained from scratch. When we finetune all layers, accuracy increases further to 94.82%. This shows that the initial features learned are useful for general purpose face-based tasks, and that the learned weights are suitable as a starting point for further adaptation. Convergence is also a lot quicker when pretrained as shown in Table 3. An additional advantage of pretraining is that it allows training with fewer images, which we show for 20 and 50 images in Table 4.

---

[2]Smith, B. M., Zhang, L., Brandt, J., Lin, Z., & Yang, J. Exemplar-based face parsing. ICCV 2013

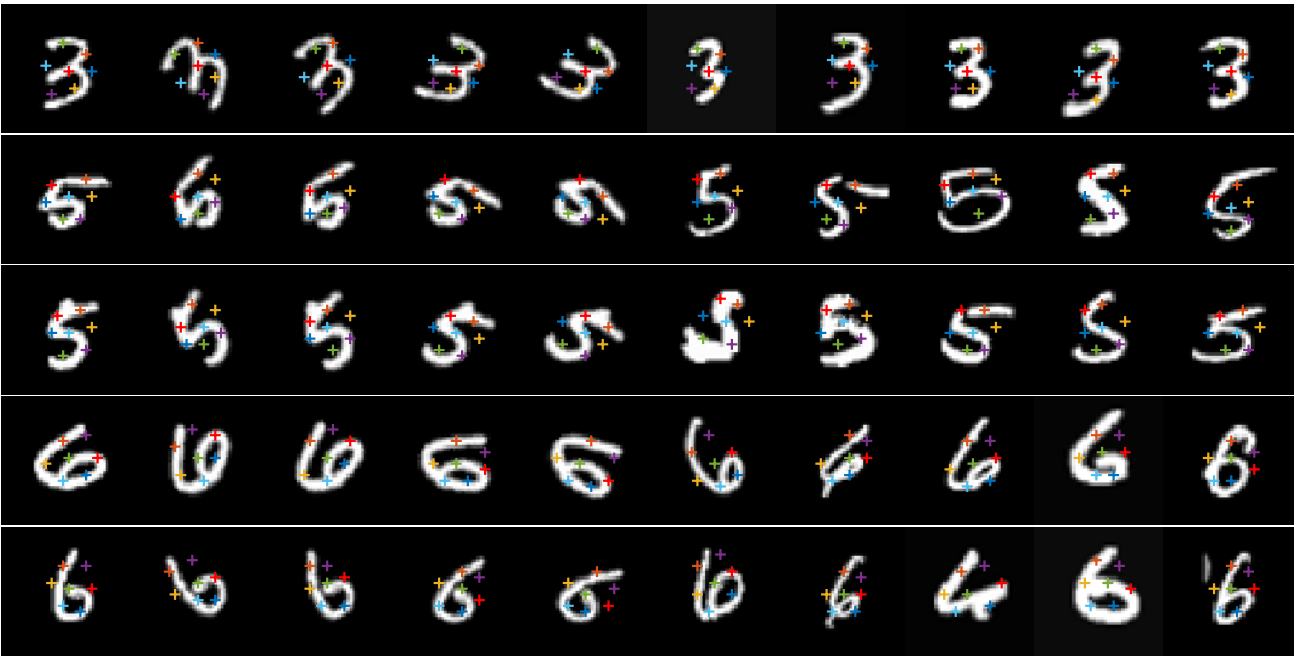

Figure 3. Three 7-landmark networks on MNIST (digits 3,5,6). The first five columns show rotations of the same instance $(0°, -50°, -30°, 30°, 50°)$ the rest show arbitrary instances.

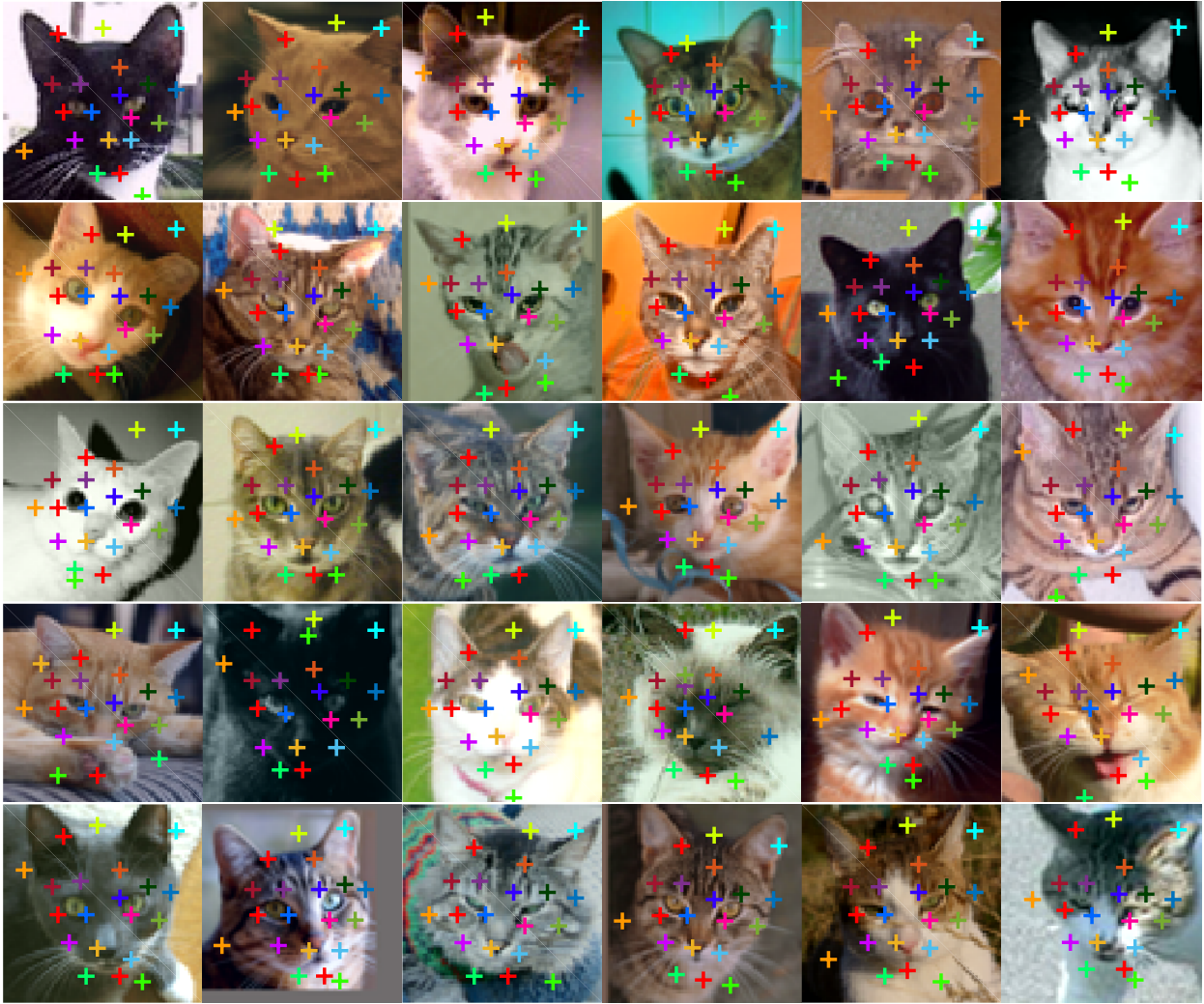

Figure 4. 20-landmark cat network

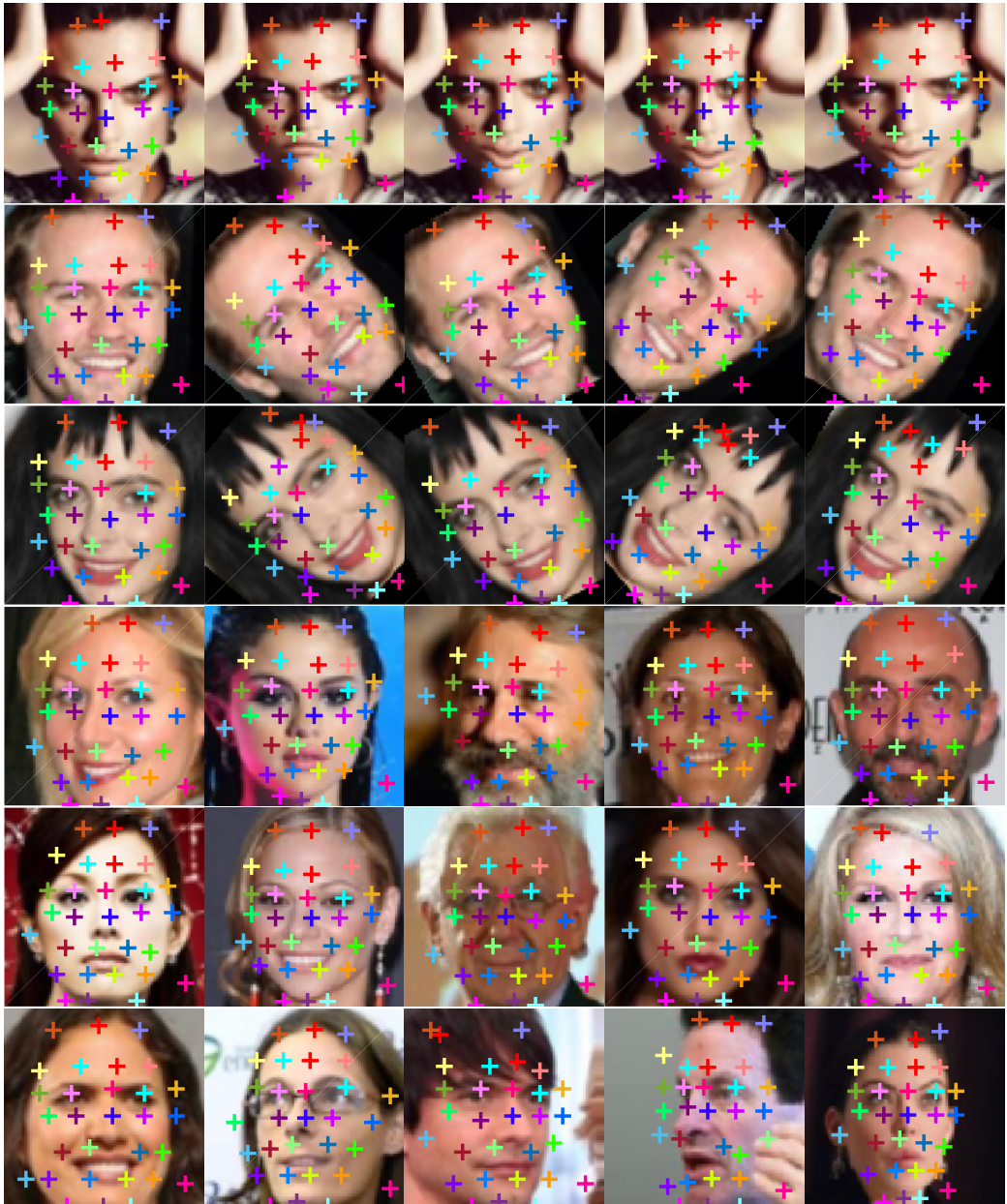

Figure 5. 30-landmark network on CelebA. Row 1: synthetic warps. Rows 2-3: rotations. Rows 4-6: arbitrary instances.

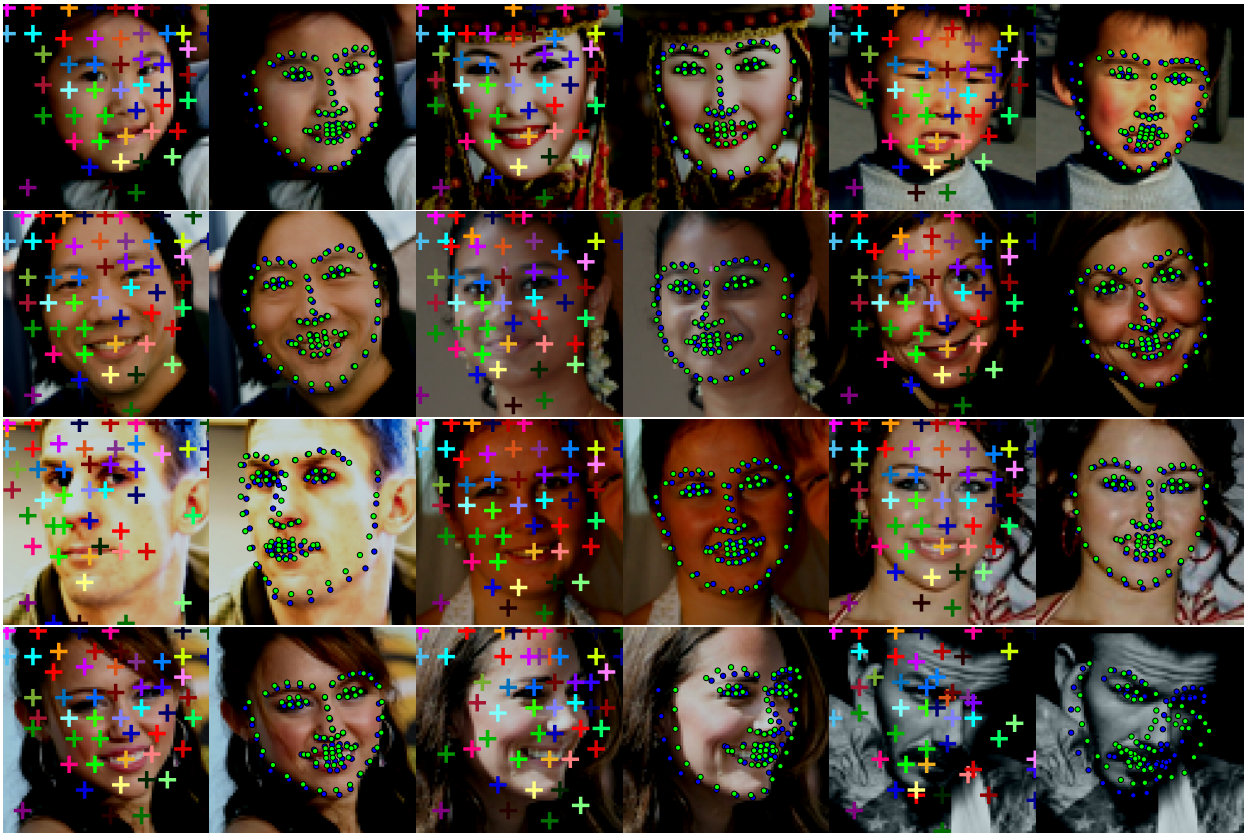

Figure 6. 30-landmark network and regressor output on 300-W. Green circles are predictions, blue circles are ground truth. The last example shows a failure case.

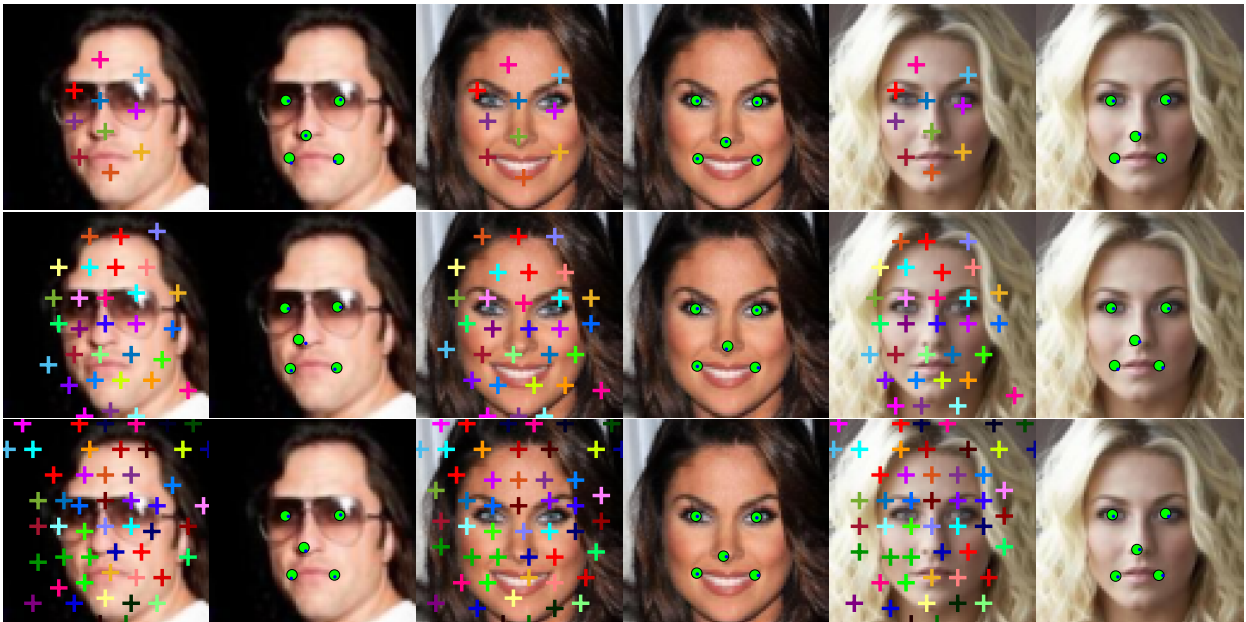

Figure 7. Unsupervised landmarks and regressor predictions for 10, 30 and 50 landmark networks in rows 1, 2 and 3 respectively. Green circles are predictions, blue circles ground truth.

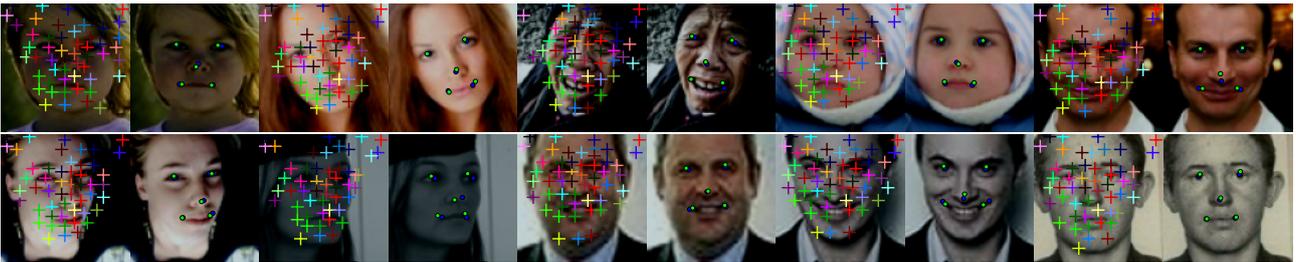

Figure 8. AFLW: Unsupervised landmarks from 51-landmark network and regressor predictions. Green circles are predictions, blue circles ground truth.